\documentclass[letterpaper, 10 pt, conference]{ieeeconf}
\usepackage{amsmath,amsfonts}
\usepackage{algpseudocode}
\usepackage{algorithm}
\usepackage{array}
\usepackage{mathtools}
\usepackage{booktabs}
\usepackage{textcomp}
\usepackage{stfloats}
\usepackage[table]{xcolor}
\usepackage{url}
\usepackage{verbatim}
\usepackage{graphicx}
\usepackage{color} 
\usepackage{drop_acronyms}
\usepackage{multirow}
\usepackage{amssymb}
\usepackage{makecell}
\newcommand{\changed}[1]{\textcolor{black}{#1}}

\usepackage{hyperref}
\hypersetup{
    colorlinks=true,
    linkcolor=red,
    filecolor=magenta,      
    urlcolor=blue,
    pdftitle={darkgs},
    pdfpagemode=FullScreen,
    }
\usepackage{algpseudocode}

\makeatletter
\let\endfigure=\end@float
\let\endtable=\end@float
\makeatletter

\usepackage[
    style=ieee,
    doi=false,
    isbn=false,
    url=false,
    eprint=false,
    backend=bibtex,
    natbib=true
    ]{biblatex}
\IEEEoverridecommandlockouts
\title{\LARGE \bf \changed{RS-ModCubes: Self-reconfigurable, Scalable, Modular Cubic Robots for Underwater Multitasking}}

\author{
    Jiaxi Zheng$^{*1,3}$, Guangmin Dai$^{*2,3}$, Botao He$^{4}$,  
    Zhaoyang Mu$^{3}$, Zhaochen Meng$^{3}$,\\Tianyi Zhang$^{1}$, Weiming Zhi$^{1, \dagger}$, Dixia Fan$^{3,5 \dagger}$
    \thanks{$^{1}$Robotics Institute, School of Computer Science, Carnegie Mellon University, Pittsburgh, PA 15213, USA.}
    \thanks{$^{2}$Zhejiang University, Hangzhou, Zhejiang 310027, China.}
    \thanks{$^{3}$Key Laboratory of Coastal Environment and Resources of Zhejiang Province, School of Engineering, Westlake University, Hangzhou, Zhejiang 310030, China.}
    \thanks{$^{4}$Department of Computer Science, University of Maryland, College Park, MD 20742, USA.}
    \thanks{$^{5}$Institute of Advanced Technology, Westlake Institute for Advanced Study, Hangzhou, Zhejiang 310024, China.}
    \thanks{$\dagger$Corresponding author. Weiming Zhi:wzhi@andrew.cmu.edu; Dixia Fan: fandixia@westlake.edu.cn.}
    \thanks{*Jiaxi Zheng and Guangmin Dai contributed equally.}
}

\begin{document}
\maketitle
\thispagestyle{empty}
\pagestyle{empty}

\begin{abstract}
\changed{This paper introduces a reconfigurable underwater robot system, RS-ModCubes, which allows scalable multi-robot configuration. An RS-ModCubes system consists of multiple ModCube modules, that can travel underwater with 6 DoFs and assemble with each other into a larger structure with onboard electromagnets. This system is designed to support diverse underwater applications through modularity and reconfigurability, eliminating the need to customize mechanical designs for a specific task.
We present a modeling framework tailored for such reconfigurable robot systems, with hydrodynamics integrated based on Monte Carlo approximation. 
A model-based feedforward \ac{PD} controller serves as the baseline for control. Inspired by dexterous manipulation, we evaluated the robot’s maximum task wrench space and power efficiency, compared against four commercial underwater robots. 
RS-ModCubes is validated via both real-world experiments and simulations, including individual and multi-module trajectory tracking and hovering docking.}
We open-source the design and code to facilitate future research: \url{https://jiaxi-zheng.github.io/ModCube.github.io/}.
\end{abstract}


\section{Introduction}
Underwater robots allow us to explore the vast benthic world, even reaching the deepest regions of the ocean, such as the Mariana Trench. However, these robots are often costly to build and limited to their designed tasks: The configuration of an underwater robot is typically specific to its designated application and does not generalize to other tasks. Such task-specific designs require extensive customization, raising costs and limiting their utility across diverse operations, which presents significant barriers to widespread adoption in research and commercial settings.

\begin{figure}[!t]
\centering
\includegraphics[width=\linewidth]{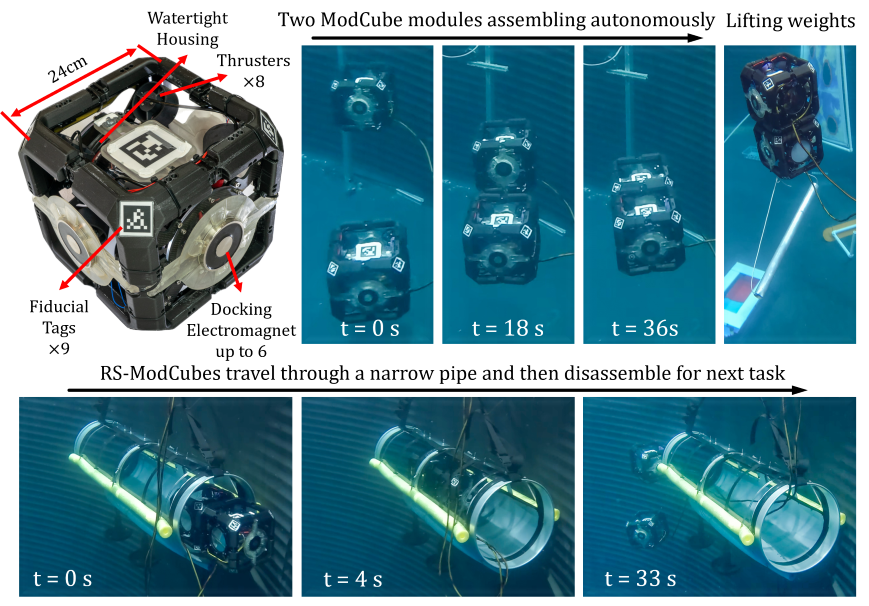}
\vspace{-20pt}
\caption{
ModCube's design (top left) features 8 thrusters allowing omnidirectional motion and docking mechanism for assemble and reconfiguration. Two ModCube modules can assemble into RS-ModCubes autonomously. 
With RS-ModCubes, we show the capability of lifting a metal pipe from the bottom of the water tank, passing through a narrow pipe, and disassembling for reconfiguration.
}
\label{pic:highbay_demo}
\end{figure}

\changed{To address these limitations, we introduce a low-cost modular robot platform, RS-ModCubes. RS-ModCubes' design features omnidirectional mobility, and multiple modules, ModCube, can be reconfigured into various configurations and shapes. In the related study, such robots are often termed as \acp{MSRR}~\cite{yim2007modular}. In this work, our objective is to take advantage of the modularity and reconfigurability of RS-ModCubes for practical applications, including underwater inspection, environmental monitoring, and search and rescue.}

\changed{ModCube robots (Section~\ref{sec:ModCube_Design}) are designed with symmetric structure to travel in harsh underwater environments with 6 \acp{DoF}. We developed a \ac{MDS} to support modular docking and operate full-body planning and navigation tasks. 
The modular design allows adaptation to different tasks that require different robot structures or under different space limitations.}

\changed{We model the floating-base system dynamics of ModCube and RS-ModCubes in Lagrangian form (Section~\ref{sec:ModCube_model_control} and~\ref{sec:RS_ModCubes_model_control}). A critical challenge lies in characterizing hydrodynamics due to morphing nature of RS-ModCubes. 
We use Monte Carlo method to approximate frontal area values and store directional drag forces in a Lookup Table for efficient retrieval. 
A model-based \ac{PD} controller is implemented for state control.}

\changed{We propose a novel characterization and visualization method (Section~\ref{sec:Morphological_Characterization}) inspired by dexterous manipulation to quantify the capabilities of underwater robots. This method evaluates task space as determined by actuator allocation, using metrics such as Willmore energy and Dirichlet energy to assess workspace continuity and smoothness. These insights guide optimal configuration selection for future complex reconfigurations.}

\begin{figure}[!t]
\centering
\includegraphics[width=0.9\linewidth]{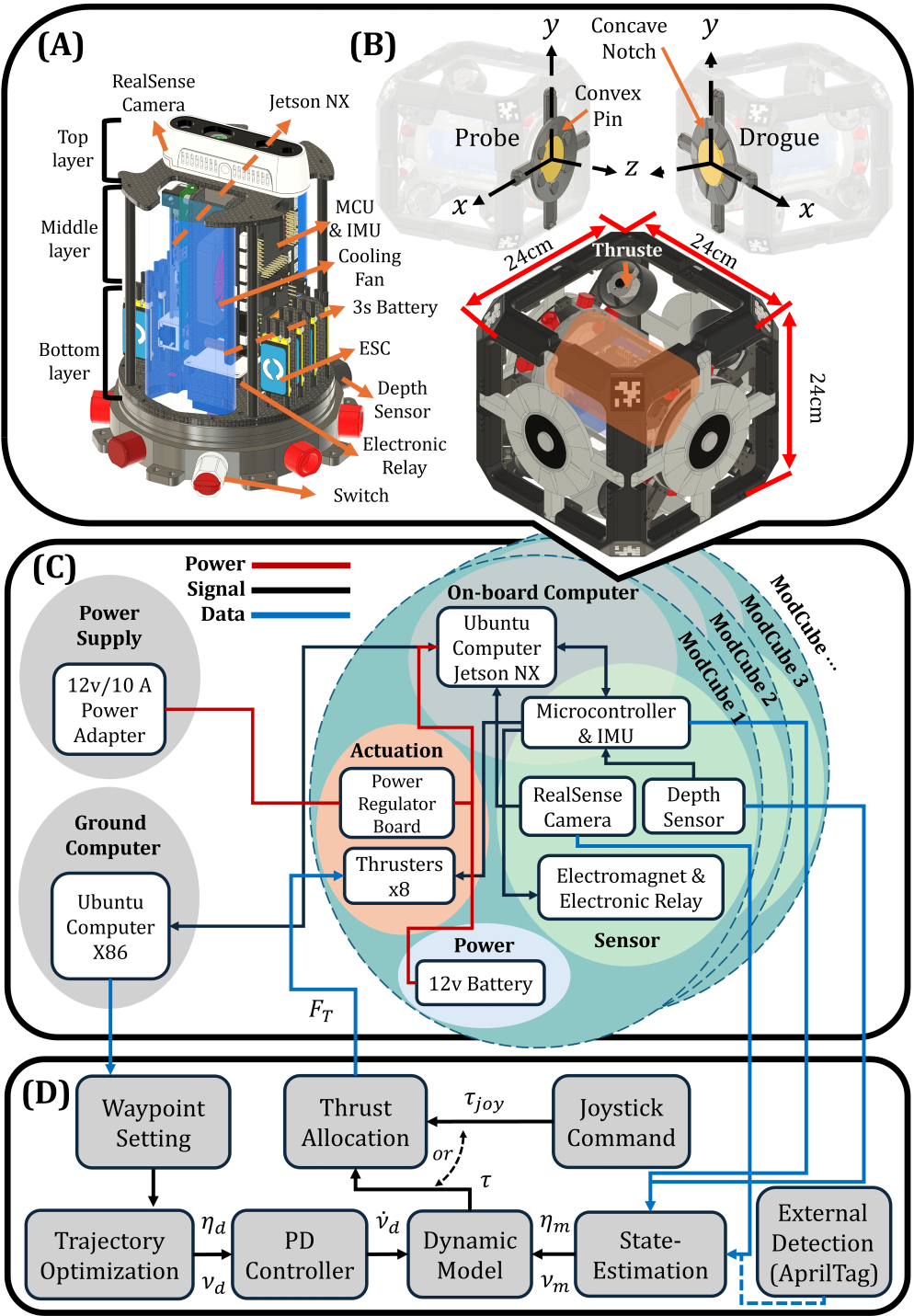}
\vspace{-7pt}
\caption{
System components and architecture. (A) Interior view of ModCube's cabin. (B) Exterior view of ModCube with the cabin highlighted in orange, and \ac{MDS} layout in the top. (C) Electronic system architecture. (D) Control and planning framework.
}
\label{pic:system_layout}
\end{figure}

\changed{This work lays a methodological groundwork for system modeling, control strategies, and the analysis of trade-offs in actuation, configuration, and power allocation, serving as a basis for future scaling efforts.
We demonstrate trajectory tracking for ModCube and RS-ModCubes through a series of real-world and simulation experiments (Section~\ref{sec:experiments}), along with reconfiguration experiments involving two ModCube modules.}

Our main contributions are summarized as follows:
\begin{enumerate}
    \item The design details of the ModCube and RS-ModCubes system.
    \changed{\item A modeling framework for individual ModCube and RS-ModCubes dynamics.
    \item A structured characterization method that quantifies and visualizes robot morphological performance.
    \item Experimental validation in water tanks, demonstrating the feasibility of the design and modeling framework in the real world.}
\end{enumerate}

\section{Related Work}
\subsection{Configuration Setup of Underwater Robots}
Underwater robot designs vary widely based on task requirements. Full-size \acp{ROV} are effective for operations but suffer from limited mobility due to their large size~\cite{yang2018grand}. Compact \acp{ROV}, typically box-shaped with five to eight thrusters, such as those in our previous work~\cite{xu2022design}~\cite{xu2023deep}, enable omnidirectional maneuverability, but exhibit poor hydrodynamic performance. For long-range missions, torpedo-shaped robots~\cite{fossum2019toward} offer streamlined efficiency.
In summary, today's underwater robots are primarily designed for a single purpose that does not generalize to different tasks. 

In this paper, we present ModCube, a modular robotic platform, and the box-shape designed for ease of manufacturing. ModCube is equipped with eight thrusters arranged in two groups of four: four thrusters are vertically inclined at 45 degrees, with a similar configuration applied horizontally. Building on our earlier work~\cite{xu2022design}, this design enables omnidirectional mobility, positioning ModCube as a versatile solution for a wide range of underwater tasks.

\subsection{Modular Robotic Swarms}
\changed{Swarm robotic systems exhibit versatility across diverse environments and applications. We categorize their capabilities into three primary types:}
\subsubsection{Shape Reconfiguration}
\changed{Shape reconfiguration involves transitioning from individual modules to a joint configuration or modifying an existing configuration. 
Ground-based systems such as 
SMORES-EP~\cite{daudelin2018integrated} and 
multi-legged robots~\cite{ozkan2021self} showcase adaptability to varying terrains. 
M-Blocks~\cite{romanishin2013m} focus on momentum-driven actuation for reconfiguration. 
In aerial domains, platforms like Modquad~\cite{saldana2018modquad} achieve stable operations through self-assembly.}
\subsubsection{Collaborative Construction}
\changed{Collaborative construction involves discrete modules working together to build larger, programmable structures. Reprogrammable robots~\cite{gregg2024ultralight}, for example, create lattice structures for adaptive infrastructure, while termite-inspired robots~\cite{werfel2014designing} enable autonomous assembly through low-level individual actions that lead to collective high-level outcomes.}

\subsubsection{Formation Tasks without Interaction}
\changed{This category encompasses swarms that operate without direct interaction between modules. Examples include distributed force control methods~\cite{wang2016force}, where modules collaborate indirectly. 
Fish-inspired swarm systems~\cite{berlinger2021implicit} focus on bio-behavioral exploration, prioritizing collective behavior over task adaptability or reconfiguration.}

\subsection{Underwater Reconfigurable Modular Robots}
\changed{Underwater Reconfigurable robots can be classified into passive and active reconfiguration systems.}
\subsubsection{Passive Reconfiguration}
\changed{Passive reconfiguration systems rely on human-in-the-loop assembly~\cite{wu2008uss}, restricting their adaptivity in complex environments. Examples include bio-inspired designs like Soft Robotic Fish (SoFi)~\cite{katzschmann2018exploration} and heterogeneous modular cubic robots~\cite{zhou2024cubic}, which lack integrated capabilities for navigation, planning, and perception.}
\subsubsection{Active Reconfiguration}
\changed{Active underwater self-reconfiguration systems are scarce due to the nascent nature of the field and the significant challenges associated with control, modeling, and conducting swarm hardware experiments underwater. 
The work~\cite{wang2024miniature} lacks advanced high-end capabilities.
An energy-heuristic strategy for self-reconfiguration was proposed in~\cite{furno2017self}, but its practical implementation remains largely unvalidated.}

\section{ModCube Design}
\label{sec:ModCube_Design}
\subsection{System Overview}
\subsubsection{Onboard Computation}
Each ModCube can \changed{optionally} integrate an Nvidia Jetson NX for onboard perception, planning, and control, enabling a full-stack robotics pipeline without tethering. \changed{Alternatively, a ground-based computer can enable these tasks via tethered communication.}

\subsubsection{Electronics Layout}
We organize electronic components into 3 layers to optimize cabin space for electronics following a bottom-up design strategy (* denote optional components): 
\begin{itemize}
    \item Bottom layer: Electric relay, Battery*, \ac{ESC},
    \item Middle layer: \ac{MCU}, Onboard Computer*, Power Distribution Board,
    \item Top layer: USB Camera*
\end{itemize}

\subsubsection{Dual Power Supply}
\changed{ModCube supports both external power supply and onboard power supply by equipping batteries, offering flexibility in operational scenarios depending on experiment or deployment requirements.}

\subsubsection{Cost Efficiency}
\changed{Each ModCube is priced at approximately \$800 (excluding the onboard computer), with customized thrusters costing \$14 each and delivering a maximum thrust of 10~N. 
This design representing a cost-effective solution compared to high-end commercial systems, as illustrated in Fig.~\ref{pic:space_metrics_comparison} D--G.}

\subsubsection{Modularity and Fabrication}
\changed{The modular design enables ModCube to accommodate up to six docking disks. For visualization, Fig.~\ref{pic:system_layout}(B) depicts a configuration with four docking disks, while the experiments in Fig.~\ref{pic:spiral_trajectory_tracking} and Fig.~\ref{pic:docking} simplify this setup by using only two disks.}

The ModCube features a dual-layer shell architecture:
\changed{The inner layer consists of a metal frame, providing structural support,}  
\changed{while the outer layer offers collision protection and serves as a modular attachment interface for components such as docking disks, manipulators, \ac{DVL}, which measures the velocity of the vehicle relative to the sea bottom, or scanning sonars.}

The outer layer and thruster cowling are fabricated from 3D-printed nylon PA12, coated with waterproof paint, allowing operation at depths of up to 50~m. Thruster propellers are 3D-printed from aluminum, and the cowlings, with a diameter of 56~mm, enclose waterproof brushless motors. The watertight housing consists of an acrylic cylindrical body with an aluminum flange for secure cable connections.

\subsection{Magnetic Docking System}
A high-tolerance docking mechanism is critical for enabling robots to dock and detach smoothly during self-assembly.  
\changed{We adopt \ac{MDS} design~\cite{branz2020miniature}, where an electronic relay controls the power state of the electromagnet, facilitating docking and detachment.  
The docking mechanism incorporates two disk types: Probe and Drogue, as shown in Fig.~\ref{pic:system_layout}(B). 
Each disk features a circular arrangement of convex pins and concave notches in an alternating pattern, visible from a side projection. 
This design ensures secure interlocking and prevents relative rotation between the Probe and Drogue disks in the docked state.  
Additionally, a guidance rail is integrated into the docking disk to provide a positioning tolerance of 4~mm and a rotational tolerance of 50~degrees. The docking disks are fabricated using Photopolymer Resin material.}

\begin{figure}[!t]
\centering
\includegraphics[width=0.9\linewidth]{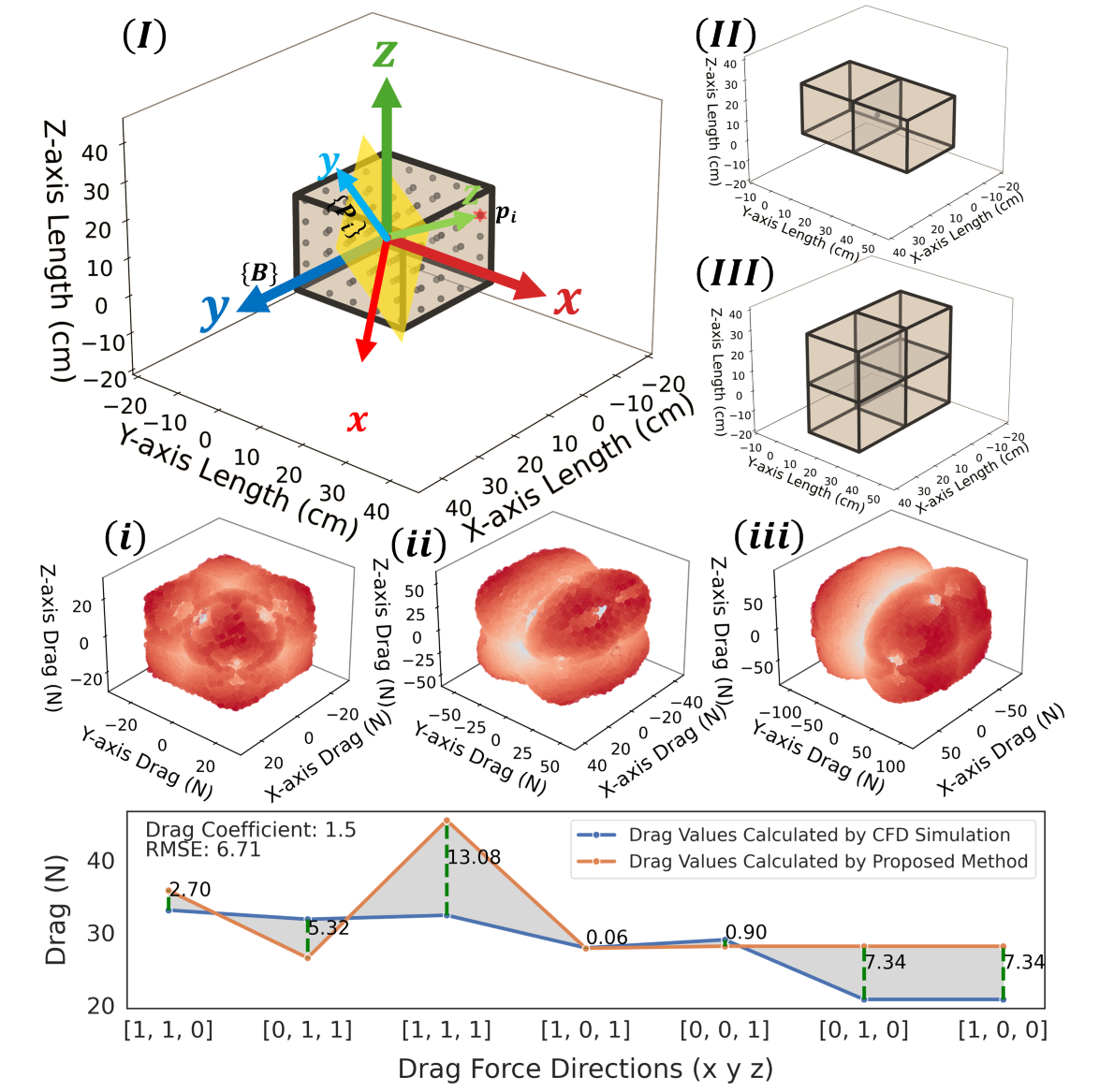}
\vspace{-8pt}
\caption{The top row (\textit{I} to \textit{III}) illustrates different RS-ModCubes configurations: single, line, and box. 
The middle row (\textit{i} to \textit{iii}) depicts the corresponding approximated drag force space 
\(\boldsymbol{D}_{lut}\) . 
The body frame $\{ \mathcal{B} \}$ is shown in \textit{I}, with an arbitrary projection plane highlighted in yellow. The random index $p_{i} \in P_{d}$ is indicated by a red star. 
The bottom row shows the comparison between approximated drag force values by the proposed method and \ac{CFD} simulation result, reaching a low \ac{RMSE} of 6.71 N.
}
\label{pic:hydrodynamics}
\end{figure}

\section{ModCube: Model and Control}
\label{sec:ModCube_model_control}
\subsection{ModCube Dynamics}
Following Fossen's formulation \cite{fossen2011handbook}, the general dynamics of underwater robots are expressed in matrix vector notation. 
\changed{Position and orientation are represented by \( \boldsymbol{\eta} \in \mathbb{R}^6 \) in World-fixed frame \(\{\mathcal{W}\}\). Velocity is represented by \( \boldsymbol{\nu} \in \mathbb{R}^6 \) in body-fixed frame $\{\mathcal{B}\}$.} 
The kinematic relationship is:
\begin{equation}
\dot{\boldsymbol{\eta}} = J(\boldsymbol{\eta}) \boldsymbol{\nu}
\end{equation}
where \(J(\boldsymbol{\eta})\) is the Jacobian matrix for the kinematic transformation.
The single ModCube's dynamic model expressed in $\{\mathcal{B}\}$ is given by:
\begin{equation}
\label{eq:singledynamics}
\mathbf M \dot{\boldsymbol{\nu}} + \mathbf C(\boldsymbol{\nu}) \boldsymbol{\nu} + \boldsymbol{D}_{lut} (\boldsymbol{\mathbf{v}_r})\boldsymbol{\mathbf{v}_r} + \mathbf g(\boldsymbol{\eta})  = \boldsymbol{\tau}
\end{equation}
where \( \mathbf M \) is the combined mass matrix including rigid body \( \mathbf M_{RB}\) and added mass effect \( \mathbf M_A\). 
\( \mathbf C(\boldsymbol{\nu}) \) accounts for Coriolis and centrifugal forces. 
\changed{\(\boldsymbol{D}_{lut} (\boldsymbol{\mathbf{v}_r})\) denotes the drag force retrieved from lookup table defined in Algorithm~\ref{algorithm:calcu}.} 
\( \mathbf g({\boldsymbol{\eta}}) \) includes buoyancy and gravitational forces designed to be zero. 
The wrench generated by the actuators is represented by \( \boldsymbol{\tau} \). \changed{\(\mathbf{v}_r\) is the relative velocity between the robot and the surrounding fluid field.}

\subsection{ModCube Control}
\changed{Model-based \ac{PD} controller with feed-forward compensation can be written as:
\begin{equation}
\label{eq:PDcontrol}
\boldsymbol{\tau} = K_P \tilde{\boldsymbol{\eta}} + K_D \dot{\tilde{\boldsymbol{\eta}}} + \mathbf M \dot{\boldsymbol \nu}_d + \mathbf C({ \boldsymbol\nu}_m){\boldsymbol \nu}_m + \boldsymbol{D}_{lut} (\boldsymbol{\mathbf{v}_r})\boldsymbol{\mathbf{v}_r}
\end{equation}
where, \(K_P\) and \(K_D\) represent the proportional and derivative gains respectively. 
\(\tilde{\boldsymbol{\eta}} \in \mathbb{R}^6\) denotes the position and orientation error. 
\({\boldsymbol {\nu}_d}\) and \({\boldsymbol \nu}_m\) are desired velocity and measured velocity in \(\{\mathcal{W}\}\).}

\begin{figure}[!t]
\centering
\includegraphics[width=\linewidth]{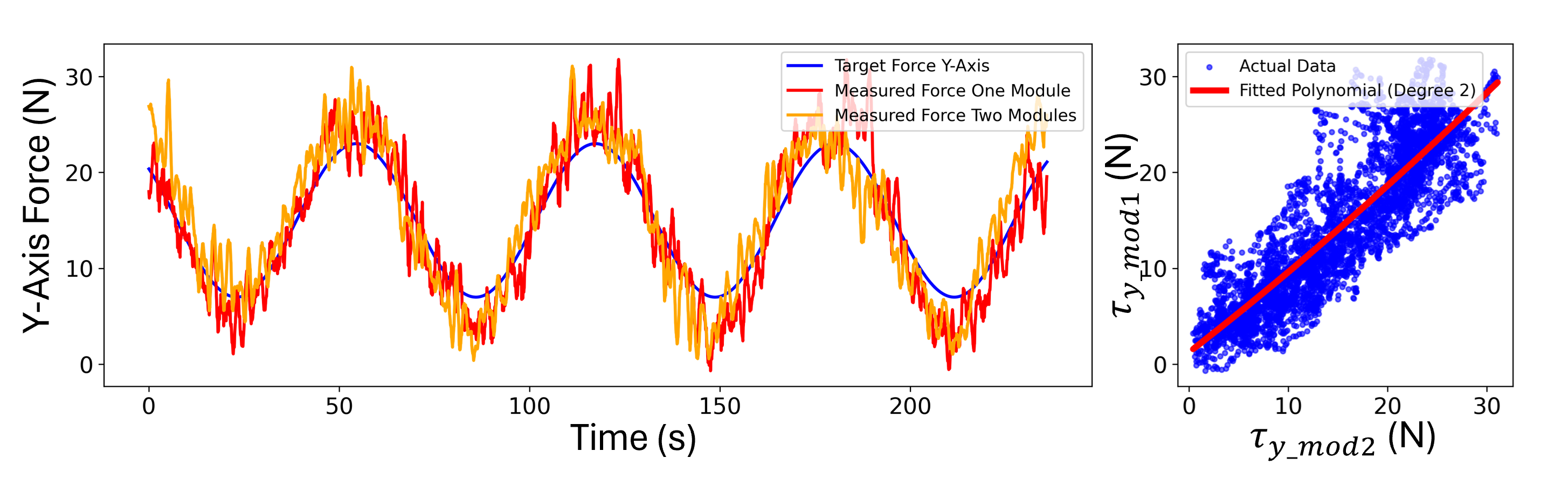}
\vspace{-23pt}
\caption{
Comparison of measured force data between an individual ModCube and a two-module RS-ModCubes during target force trajectory tracking (left). A polynomial regression fit for the measured forces (right).
}
\label{force_tracking}
\end{figure}

\subsection{Thrust Allocation}
Relationship between control input $u$ and power consumption $P$ of thruster is modeled using polynomial functions:
\begin{equation}
\label{eq:thrust_model}
P(f) = c_f \boldsymbol{\zeta}(f(u)), \quad 
f(u) = c_u \boldsymbol{\zeta}(u)
\end{equation}
where, \(f(u)\)is the thrust force. \changed{$\boldsymbol{\zeta}(x) = [1, x, x^2, \ldots, x^5]^T $ }represents the vector of polynomial basis functions, while \(c_f\) and \(c_u\) denote the polynomial parameter matrices.
The overall force and moment exerted on the vehicle are calculated as follows:
\begin{equation}
{
\label{eq:Jacobian}
{J_t} = \sum_{i=1}^{n_t} \left[ \begin{matrix}
p_i^t \\ 
r_i^t \times p_i^t 
\end{matrix} \right], \quad F_T = {J_t}^+ \boldsymbol{\tau}
}\end{equation}
where \(\boldsymbol{\tau} \in \mathbb{R}^{6}\) is the applied wrench from Eq.~\ref{eq:singledynamics}, \({\quad F_T} \in \mathbb{R}^{n_t \times 1}\) is the thrust matrix (\(n_t\) is the total number of the thruster), and \({J_t}\) is Jacobian matrix mapping joint space to work space, \({J_t}^{+}\) is the pseudo-inverse of \({J_t}\). The position vector \( p_i^t \) and rotation vector \(r_i^t\) are defined relative to the vehicle's \ac{CoM}. 

\changed{To evaluate power efficiency with given unit wrench space. 
The corresponding power consumption space is denoted as
\(\Omega_{P}=\bigoplus_{i=1}^{n_s}\{P P_d \in \mathbb{R}^6 \mid P/\|P\|=P_d/\|P_d\|,\{\mathcal{B}\}\}\), definition of \(P_d\) in Algorithm~\ref{algorithm:calcu}~(line 3--5). 
Results are detailed in Fig.~\ref{pic:space_metrics_comparison} and Table~\ref{tab:space_metrics_comparison}. The \(\bigoplus\) denotes Minkowski sum results in the sets of parameter space.}

\subsection{ModCube Hydrodynamics} 
\changed{In fluid environments, the shape of locomotors vary over time, leading to variations in drag forces~\cite{gatta2015planimetric}. Accurately modeling these dynamic changes is crucial for understanding and predicting the behavior of shape-varying.
The drag force \( {F}_{\text{drag}} \in \mathbb{R}^6 \) is given by:}
\vspace{-7pt}
\begin{equation}
\label{eq:nasadrag}
{F}_{\text{drag}}(\mathbf{v}_r) = -\frac{1}{2} \rho C_d A \| \mathbf{v}_r \| \mathbf{v}_r,
\end{equation}
\vspace{-12pt}

where \( C_d \) is the drag coefficient, computed via \ac{CFD} in Ansys Fluent with \( \{\|\mathbf{v}_r\| = 1 \, \text{m/s} \mid \{\mathcal{B}\} \}\), and \( \rho \) is the fluid density. The frontal area \( A \), orthogonal to the velocity vector, is approximated in Algorithm~\ref{algorithm:calcu}.
\( \mathbf {v}_r \) denotes the relative velocity of the robot with respect to the fluid, ensuring \(\{{F}_{\text{drag}} / \|{F}_{\text{drag}}\| = \mathbf{v}_r /\|\mathbf{v}_r\|\}\). 

\changed{The mid graph in Fig.~\ref{pic:hydrodynamics} visualizes the approximated drag spaces \(\boldsymbol{D}_{lut}\) for three distinct RS-ModCubes configurations. 
Calculated drag forces \({F}_{\text{drag}}\) are compared with \ac{CFD} results for an individual ModCube, shown in Fig.~\ref{pic:hydrodynamics}~\textit{(I)} in the bottom graph. The comparison spans seven flow directions, indicated on the x-axis. 
Notable deviations in the \textit{(1,1,1)} direction are attributed to sharper leading edges, which streamline the flow, reduce turbulent wake formation, and consequently lower drag forces.}

\begin{algorithm}[t]
\small
\caption{\small Drag Force Lookup Table Calculation}
\label{algorithm:calcu}
\begin{algorithmic}[1]
\State \textbf{Input:} 
\changed{\( P_s \): Set of sample points inside each ModCube as Fig.~\ref{pic:hydrodynamics}~\textit{(I)}. 
\( n_s \): Number of sample directions. 
\( \rho \): Water density. 
\( C_d \): Drag coefficient. 
\(\mathbf{v}_r\): Relative velocity}
\State \textbf{Output:} \changed{${\boldsymbol{D}_{lut}} \in \mathbb{R}^{6}$}
\For{$i = 0$ to $n_s-1$}
    \State $\varphi_i = \arccos\left(1 - \frac{2i}{n_s}\right),\theta_i = \pi \left(1 + \sqrt{5}\right) i$
    \State $P_d \leftarrow ( \cos(\theta_i) \sin(\varphi_i), \sin(\theta_i) \sin(\varphi_i), \cos(\varphi_i))^\top$
\EndFor
\For{ $p_i \in P_d, \{i \in n_s\}$}
    \State $\mathbf{e}_z = [0, 0, 1]^\top$, ${v}_i = {p}_i \times \mathbf{e}_z$, $c_i = {{p}^\top_i} \mathbf{e}_z$, $s_i = \|{v}_i\| \neq 0$
    \State $\mathbf{R}_i = \mathbf{I}_{3 \times 3} + {K(v)}_i + {K(v)}_i^2 \left( \frac{1 - c_i}{s_i^2} \right),{K}^\top = -{K}$
    \changed{\State \( P_{\text{proj}} = \{ (x, y) \mid (x, y, z) \in (\mathbf{R}_i P_s) \} \)
    \State $A_i = \texttt{$\alpha shape$} (P_{\text{proj}})$
    \State \( {F_{drag,i}(p_i)} = -\frac{1}{2} \rho C_d A_i {({\|p_i\|} {p}_i})^2\)
    \State \( {\tau_{drag,i}(p_i)} = -\frac{1}{2} \rho C_d A_i {({\|p_i\|} {p}_i})^{\frac{5}{3}}\)~\cite{chemel2020tartan}}
\EndFor
\State \Return \changed{$ {\boldsymbol{D}_{lut}} = \bigoplus_{i=1}^{n_s} {\{F_{drag,i}(p_i)},{\tau_{drag,i}(p_i)}\}$}
\end{algorithmic}
\end{algorithm}

\subsection{Monte Carlo Approximation for Drag Force}
\label{sec:Monte_Carlo}
\changed{The lookup table \(\boldsymbol{D}_{lut}\) is employed to retrieve direction-dependent drag forces, as detailed in Algorithm~\ref{algorithm:calcu}.}

\changed{(1) We uniformly sample points \( P_s \in \mathbb{R}^3 \) within ModCube module's body, as Fig.~\ref{pic:hydrodynamics} \textit{I}. 
These points serve as volumetric particles for approximating the area volume.}

\changed{(2) We define a set of direction vectors 
\( P_d = \bigoplus_{i=1}^{n_s} \{ p_i \in \mathbb{R}^3 \mid \|p_i\|_2 = 1\} \), which are uniformly distributed over a unit sphere. For each direction vector \( p_i \), a orthogonal projection plane \( \mathcal{P}_i \) is defined. The plane is visualized in yellow in Fig.~\ref{pic:hydrodynamics}~(\textit{I}), are specified in Algorithm~\ref{algorithm:calcu}~(lines 3--5).}

\changed{(3) We obtain a set of 2-dimensional projected point clouds by projecting \( P_s \) onto plane \( \mathcal{P}_i \). The area of projected point cloud is approximated using the \(\alpha\)-shape tool, yielding \( A_i \), as described in lines 7--11 of Algorithm~\ref{algorithm:calcu}.}

\changed{Finally, incorporating \( A_i \) into Eq.~\ref{eq:nasadrag} yields the resulting drag force.}

\section{RS-ModCubes: Model and Control}
\label{sec:RS_ModCubes_model_control}
\subsection{RS-ModCubes Connectivity Analysis}
\changed{Two conditions are used to determine ifdocking between ModCube modules is successful. If both conditions are satisfied, the docking is considered successful: }

(1) Condition 1: the current over the electromagnet exceeds a threshold of \(\pm 70 \, \text{mA}\). The threshold is suggested by the data collected from over ten real-robot docking experiments, as discussed in Section~\ref{subsec_assembly} and visualized in Fig.~\ref{pic:docking}. 
(2) Condition 2: the relative position between two modules equals the length of one ModCube and stabilizes.

\subsection{Proximity Effects of the Docking Process}
\label{ProximityEffects}
\changed{We assume docked RS-ModCubes introduce jet interference, potentially affecting system dynamics. 
To validate this hypothesis, we conducted experiments to measure the one-directional force tracking for an individual ModCube and a two-module RS-ModCubes configuration, as shown in Fig.~\ref{force_tracking}. 
The percentage difference between two tracking results' \ac{RMSE} is 0.46\%, thus we consider the interference caused by docking to be negligible.}
\subsection{RS-ModCubes Modeling}
\changed{The mass distribution and inertia of each add-on module in RS-ModCubes vary significantly, introducing notable changes to the system dynamics.
The total mass \( \mathbf{M}_{rsm} \) and position of \ac{CoM} \( \mathbf{r_{com}} \) for RS-ModCubes are computed as:}
\vspace{-5pt}
\begin{equation}
\label{eq:CoM}
\changed{\mathbf{M}_{rsm} = \sum_{i=1}^N \mathbf{m}_i, \quad
\mathbf{r_{com}} = \frac{\sum_{i=1}^N \mathbf{m}_i \mathbf{p}_i}{\mathbf{M}_{rsm}}, \quad
\mathbf{r}_i = \mathbf{p}_i - \mathbf{r_{com}}}
\end{equation}
\vspace{-13pt}

\changed{where \( \mathbf{p}_i\) and \(\mathbf{m}_i\) are position vector and mass of the \( i \)-th ModCube module, respectively. 
\(N\) is the number of ModCube module. 
\(\mathbf{r}_i\) is the relative position of each ModCube module with respect to \ac{CoM} of RS-ModCubes.
Applying the parallel axis theorem, the updated total inertia tensor \( \mathbf{I}_{rsm} \) of RS-ModCubes is:
\vspace{-7pt}
\begin{equation}
\mathbf{I}_{rsm} = \sum_{i=1}^N \left( \mathbf{I}_i + \mathbf{m}_i \left( \|\mathbf{r}_i\|^2 \mathbf{I}_3 - \mathbf{r}_i \mathbf{r}_i^\top \right) \right)
\end{equation}
\vspace{-7pt}}

\changed{where, \(\mathbf{I}_i \) is the inertia tensor of the \( i \)-th ModCube module about its own \ac{CoM}, and \( \mathbf{I}_3 \) is the \( 3 \times 3 \) identity matrix.}

\changed{The total mass matrix \( \mathbf{M}_{rsm} \) and the coriolis and centripetal matrix \( \mathbf{C}_{rsm}(\boldsymbol{\nu}) \) for RS-ModCubes are then given by:
\vspace{-7pt}
\begin{equation}
{
\mathbf{M}_{rsm} = \begin{bmatrix}
\mathbf{M}_{rsm} \mathbf{I}_3 & - \mathbf{M}_{rsm} {\mathbf{r}_i^\times}^\top \\
\mathbf{M}_{rsm} {\mathbf{r}_i^\times} & \mathbf{I}_{rsm}
\end{bmatrix} + \mathbf{M}_A
}\end{equation}
\vspace{-10pt}
\begin{equation}
\mathbf{C}_{rsm}(\mathbf{v},\omega) = \begin{bmatrix}
\mathbf{0}_{3 \times 3}& -\mathbf{M}_{rsm} \boldsymbol{\mathbf{v}}^\times  - {\mathbf{M}_A \mathbf{v}}^\times  \\
\mathbf{M}_{rsm} \boldsymbol{\mathbf{v}}^\times +  {\mathbf{M}_A \mathbf{v}}^\times & {(\mathbf{I}_{rsm} \omega)}^\times + {(\mathbf{M}_A \omega)}^\times
\end{bmatrix}
\end{equation}
where, parameters marked with\(\{^\times\}\) represent the skew-symmetric matrix form of the corresponding vector. \(\mathbf{M}_A\) denotes the added mass matrix, which is designed to be zero.}

\changed{Finally, the dynamics of RS-ModCubes are described by:
\vspace{-7pt}
\begin{equation}
\mathbf{M}_{rsm} \dot{\boldsymbol{\nu}} + \mathbf{C}_{rsm}(\boldsymbol{\nu}) \boldsymbol{\nu} + \boldsymbol{D}_{rsm,lut} (\boldsymbol{\nu})\boldsymbol{\nu} = 
\boldsymbol{\tau}
\end{equation}}
\changed{The control law for RS-ModCubes is same to Eq.~\ref{eq:PDcontrol}, giving as:
\vspace{-13pt}
\begin{align}
\boldsymbol{\tau} &= K_P \tilde{\boldsymbol{\eta}} + K_D \dot{\tilde{\boldsymbol{\eta}}} 
+ \mathbf{M}_{rsm} \dot{\boldsymbol{\nu}}_m + \mathbf{C}_{rsm}({\boldsymbol\nu}_m) \boldsymbol{\nu}_m \notag \\
&\quad + \boldsymbol{D}_{rsm,lut}(\boldsymbol{\nu}_m)\boldsymbol{\nu}_m
\end{align}}
\vspace{-16pt}

\begin{figure*}[!t]
\centering
\includegraphics[width=\linewidth]{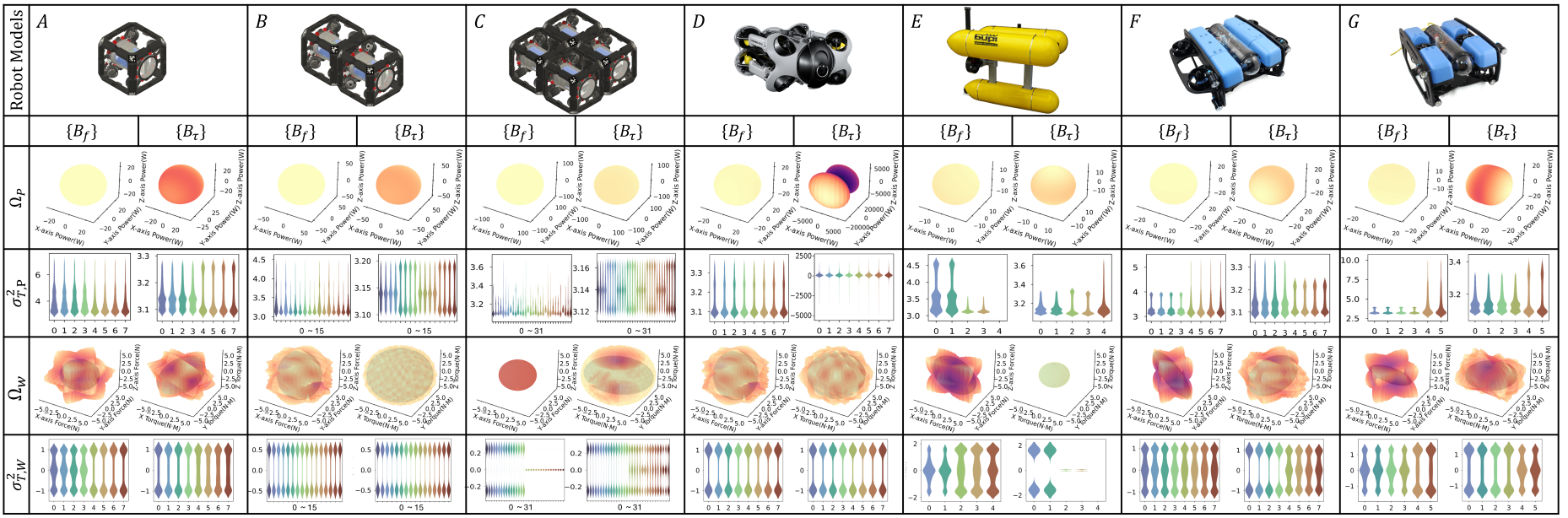}
\vspace{-16pt}
\caption{
Benchmark comparison of three differemt RS-ModCubes configurations (A--C) and four commercial underwater robots (D--G). 
The evaluation includes the power consumption space (\(\Omega_{P}\)) and the reachable wrench space (\(\Omega_{W}\)) in the body force frame (\(\{\mathcal{B}_f\}\)) and body torque frame (\(\{\mathcal{B}_{\tau}\}\)), respectively. 
Additionally, the thruster thrust distribution is analyzed using violin plots of variance metrics (\(\sigma_{T,P}^2\) and \(\sigma_{T,W}^2\)), which quantify the thrust effort distribution across force and torque components. 
X-axis of violin plots represents the thruster index, while the Y-axis indicates the distributions of thrust forces.
Detailed evaluations values are provided in Table~\ref{tab:space_metrics_comparison}.
}
\label{pic:space_metrics_comparison}
\end{figure*}

\begin{table*}[!t]
\centering
\vspace{-5pt}
\caption{Capability Comparison of Seven Different Underwater Robots}
\vspace{-5pt}
\label{tab:space_metrics_comparison}
\setlength{\tabcolsep}{4pt} 
\renewcommand{\arraystretch}{1} 
\small 
\begin{tabular}{llp{1.8cm}p{1.8cm}p{1.8cm}p{1.8cm}p{1.8cm}p{1.8cm}p{1.8cm}}
\hline
\multicolumn{2}{l}{} 
& A: \makecell{Single\\ModCube} & B: \makecell{Double\\ModCube} & C: \makecell{Triple\\ModCube} & D: \makecell{Chasing\\M2} & E: \makecell{Girona\\500} & F: \makecell{BlueROV2\\Heavy} & G: \makecell{BlueROV2} \\ \hline
\multicolumn{2}{l}{Price (USD)$\downarrow$} 
& \cellcolor{red!25}\makecell{800 }& \cellcolor{orange!25}\makecell{1600} & \cellcolor{yellow!25}\makecell{2400} & \makecell{2500} & \makecell{7000+} & \makecell{4500} & \makecell{6000} \\ \hline
\multicolumn{2}{l}{Size (cm) $\leftrightarrow$}  
& \cellcolor{red!25}\makecell{21x21x21} & \cellcolor{yellow!25}\makecell{43x21x21} & \makecell{43x43x21} & \cellcolor{orange!25}\makecell{38x26.7x16.5} & \makecell{150x100x100} & {45.7x33.8x25.4} & {45.7x43.6x25.4} \\ \hline
\multicolumn{2}{l}{Depth (m) $\uparrow$}  
& \makecell{50} & \makecell{50} & \makecell{50} & \cellcolor{orange!25}\makecell{100} & \cellcolor{red!25}\makecell{500} & \cellcolor{orange!25}\makecell{100} & \cellcolor{orange!25}\makecell{100} \\ \hline
\multirow{4}{*}{\makecell[{{}}]{\(\Omega_{P}\) \\ \(\{\mathcal{B}_f\}\) }}
& $E_{\mathcal{D}} \downarrow$ 
& \cellcolor{orange!25}$1.896e+00$ & $6.686e+00$ & $2.648e+01$ & $6.594e+00$ & \cellcolor{yellow!25}$2.932e+00$ & $3.196e+00$ & \cellcolor{red!25}$9.988e-01$ \\
& ${E_{\mathcal{W}} \downarrow}$ 
& $1.540e+11$ & \cellcolor{orange!25}$1.440e+11$ & \cellcolor{red!25}$1.410e+11$ & \cellcolor{yellow!25}$1.460e+11$ & $2.130e+13$ & $3.080e+13$ & $1.460e+13$ \\
& $P_W \downarrow$ 
& \cellcolor{yellow!25}$2.523e+05$ & $5.029e+05$ & $1.005e+06$ & $2.533e+05$ & \cellcolor{red!25}$1.586e+05$ & \cellcolor{yellow!25}$2.523e+05$ & \cellcolor{orange!25}$1.899e+05$ \\
& $\sigma_{T,P}^2 \downarrow$ 
& $3.209e+01$ & \cellcolor{orange!25}$7.688e+00$ & \cellcolor{red!25}$1.901e+00$ & $6.025e+01$ & $6.433e+01$ & \cellcolor{yellow!25}$2.616e+01$ & $5.207e+01$\\ \hline
\multirow{3}{*}{\makecell[{{}}]{\(\Omega_{W}\) \\ \(\{\mathcal{B}_f\}\) }}
& $E_{\mathcal{D}} \downarrow$ 
& \cellcolor{orange!25}$3.207e-01$ & $4.637e+01$ & \cellcolor{red!25}$7.139e-02$ & $4.637e+01$ & $6.892e+02$ & $5.303e+02$ & \cellcolor{yellow!25}$9.834e-01$ \\
& $E_{\mathcal{W}} \downarrow$ 
& $4.250e+17$ & \cellcolor{yellow!25}$6.220e+14$ & \cellcolor{red!25}$5.747e+10$ & \cellcolor{orange!25}$6.070e+14$ & $2.370e+16$ & $1.050e+17$ & $9.430e+16$ \\ 
& $V_{MIE} \uparrow$ 
& \cellcolor{orange!25}$6.801e+01$ & \cellcolor{red!25}$1.559e+02$ & \cellcolor{yellow!25}$6.259e+01$ & \cellcolor{red!25}$1.559e+02$ & $3.194e+01$ & $3.053e+01$ & $4.006e+01$\\ \hline
\multirow{4}{*}{\makecell[{{}}]{\(\Omega_{P}\) \\ $\{\mathcal{B}_{\tau}\}$}}
& $E_{\mathcal{D}} \downarrow$ 
& $3.673e+03$ & $1.565e+03$ & \cellcolor{orange!25}$1.629e+02$ & $7.644e+09$ & \cellcolor{red!25}$1.512e+02$ & \cellcolor{yellow!25}$4.667e+02$ & $5.082e+03$ \\
& ${E_{\mathcal{W}} \downarrow}$ 
& \cellcolor{orange!25}$1.570e+11$ & $2.090e+11$ & $2.240e+11$ & $3.630e+16$ & $2.540e+11$ & \cellcolor{red!25}$1.540e+11$ & \cellcolor{yellow!25}$1.740e+11$\\
& $P_W \downarrow$ 
& $3.256e+05$ & $5.336e+05$ & $1.014e+06$ & $2.761e+07$ & \cellcolor{red!25}$1.680e+05$ & \cellcolor{yellow!25}$2.755e+05$ & \cellcolor{orange!25}$2.249e+05$\\
& $\sigma_{T,W}^2 \downarrow$ 
& $7.536e+03$ & \cellcolor{orange!25}$5.652e+02$ & \cellcolor{red!25}$6.708e+01$ & $4.693e+09$ & \cellcolor{yellow!25}$1.341e+03$ & $1.622e+03$ & $1.016e+04$\\ \hline
\multirow{3}{*}{\makecell[{{}}]{\(\Omega_{W}\) \\ $\{\mathcal{B}_{\tau}\}$}}
& $E_{\mathcal{D}} \downarrow$ 
& $1.103e+01$ & \cellcolor{yellow!25}$3.757e-01$ & $3.803e+02$ & \cellcolor{orange!25}$3.981e-01$ & \cellcolor{red!25}$7.845e-02$ & $3.757e+01$ & $3.397e+02$\\
& $E_{\mathcal{W}} \downarrow$ 
& \cellcolor{orange!25}$2.760e+13$ & $6.950e+16$ & $5.830e+16$ & \cellcolor{yellow!25}$4.320e+14$ & \cellcolor{red!25}$6.612e+09$ & $2.350e+16$ & $4.690e+16$\\ 
& $V_{MIE} \uparrow$ 
& $9.983e+01$ & \cellcolor{red!25}$3.475e+02$ & \cellcolor{yellow!25}$1.830e+02$ & \cellcolor{orange!25}$2.242e+02$ & $3.478e+01$ & $1.323e+02$ & $7.718e+01$\\ \hline
\end{tabular}
\end{table*}

\section{Evaluation of Actuation Capability and Power Efficiency}
\changed{We use analytical idea from dexterous manipulation, including parameter spaces, and task polytopes~\cite{morton2024task}.}
\subsection{Reachable Wrench Space}

\changed{To evaluate the robot's capability to generate wrenches in various directions, we formulate a convex optimization problem to determine the reachable wrench space.
\(\Omega_{W} = \bigoplus_{i=1}^{n_s} \{(\mathbf{\tau}_n = \lambda P_d) \in \mathbb{R}^{6} \mid \mathbf{\lambda} \geq 0, \{\mathcal{B}\} \}\), \(P_d\) and \(n_s\) defined in Algorithm~\ref{algorithm:calcu}~(line 3--5).
\(\mathbf{\tau}_n\) is the net wrench. 
The optimization problem defined as:
\begin{equation}
\begin{aligned}\max & \quad \lambda_j, \quad j \in n_s \\
\text{subject to} & \quad {J_t}\,{F_T} = \lambda_j P_d, \quad Eq.\ref{eq:Jacobian},\\
& F_{t,\min} \leq F_{t,i} \leq F_{t,\max} \mid F_{t,i} \in F_T, i \in n_t \\
\end{aligned}
\label{eq:wrench_optimization}
\end{equation}
\vspace{-7pt}}

\changed{where, the thrust matrix ${F_T}$ and thruster number $n_t$ from Eq.~\ref{eq:Jacobian}.
The reachable wrench spaces of seven different vehicles are shown in Fig.~\ref{pic:space_metrics_comparison}~(second to last line). 
Additionally, the volume of the Maximum Inscribed Ellipsoid (MIE), denoted as \(V_{MIE}\), illustrates the maximum stable wrench task space~\cite{morton2024task}. The corresponding \(V_{MIE}\) values are provided in Table~\ref{tab:space_metrics_comparison}, and visualized inside wrench space.}

\subsection{Morphological Characterization}
\label{sec:Morphological_Characterization}
\changed{To evaluate the parameter spaces \(\Omega_{W}\) and \(\Omega_{P}\), we analyze the relative Willmore energy (\(E_{\mathcal{W}}\)) and Dirichlet energy (\(E_{D}\)), which offer insights into smoothness, continuity, and transition stability. 
Additionally, the variance in thrusters' effort serves as a key metric for assessing individual lifespan.}

\changed{\(E_{\mathcal{W}}\) is calculated by:
\vspace{-7pt}
\begin{equation}
E_{\mathcal{W}}({\Omega}) = \int_{{\Omega}} H^2 \, dA - \int_{{\Omega}} K \, da
\end{equation}
\vspace{-10pt}}

\changed{\(H\) denotes the mean curvature at a vertex sample point \(\{ {p}_i \in P_d \mid i \in n_s\}\), where the set of points is derived from Algorithm~\ref{algorithm:calcu}. 
Here, \(K\) represents the Gaussian curvature at \(p_i\), and \(a\) is the differential surface area element associated with \(p_i\).}

\changed{\(E_{\mathcal{D}}\) is calculated as: 
\vspace{-7pt}
\begin{equation}
\text{\small $E_{\mathcal{D}} = \sum_{l=0}^{l_{\text{max}}} \sum_{m=-l}^{l} l(l+1) \left| \arg\min_{a} \| Y a - r \|_2 \right|^2$}
\end{equation}
\vspace{-9pt}}

\changed{where \(r\) is the radial distance of the vertices from the centroid in spherical coordinates, and \(Y\) is the matrix of spherical harmonic basis functions \(Y_l^m(\theta, \phi)\) representing the surface geometry. The coefficients \(a\) are obtained by solving the least-squares problem \(\arg\min_{a} \| Y a - r \|_2\). \(l_{\text{max}} = 10\) is the maximum harmonic degree, and \(l(l+1)\) is a weighting factor that penalizes higher-order harmonics to prioritize surface smoothness.}

\subsection{Performance Evaluation}
\changed{We compare seven underwater robots using the proposed metrics, each robot characterized by a rigid fixed-frame design and full autonomy capabilities, including perception and navigation. The robots vary in thruster count, ranging from 5 to 32, and exhibit diverse joint space configurations.
Prices are normalized by excluding additional sensors and retaining only essential components, including thrusters, cameras, \ac{MCU}, depth sensors, and necessary electronic systems. 
Onboard computers are not included in the cost analysis.}

\changed{(1) We evaluate each robot's power consumption under a unit wrench applied in uniformly distributed directions with number (\(n_s\)) over a unit sphere, as described in Algorithm~\ref{algorithm:calcu}~(lines 3--5). The resulting thrust power distribution is visualized as \(\Omega_{P}\) in Fig.~\ref{pic:space_metrics_comparison}.
(2) We calculate the reachable wrench space using optimization method defined in Eq.~\ref{eq:wrench_optimization}. Each robot's power is normalized, and the corresponding reachable wrench space is shown as \(\Omega_{W}\) in Fig.~\ref{pic:space_metrics_comparison}. The summarized results are presented in Table~\ref{tab:space_metrics_comparison}.}

\changed{Triple ModCube achieves the lowest \(\sigma_{T}^2\), suggesting theoretically longest thruster lifespan, while Double ModCube has the highest inscribed ellipsoid volume \(V_{MIE}\) of wrench space, representing the largest wrench task space. 
This highlights the system's actuation capability, robustness and stability are modifiable by reconfiguration.}

\changed{The Girona 500 performs well across several metrics. However, its five-thruster configuration limits operational wrench reachability and agility despite its notable smoothness.}

\begin{figure}[!t]
\centering
\includegraphics[width=0.9\linewidth]{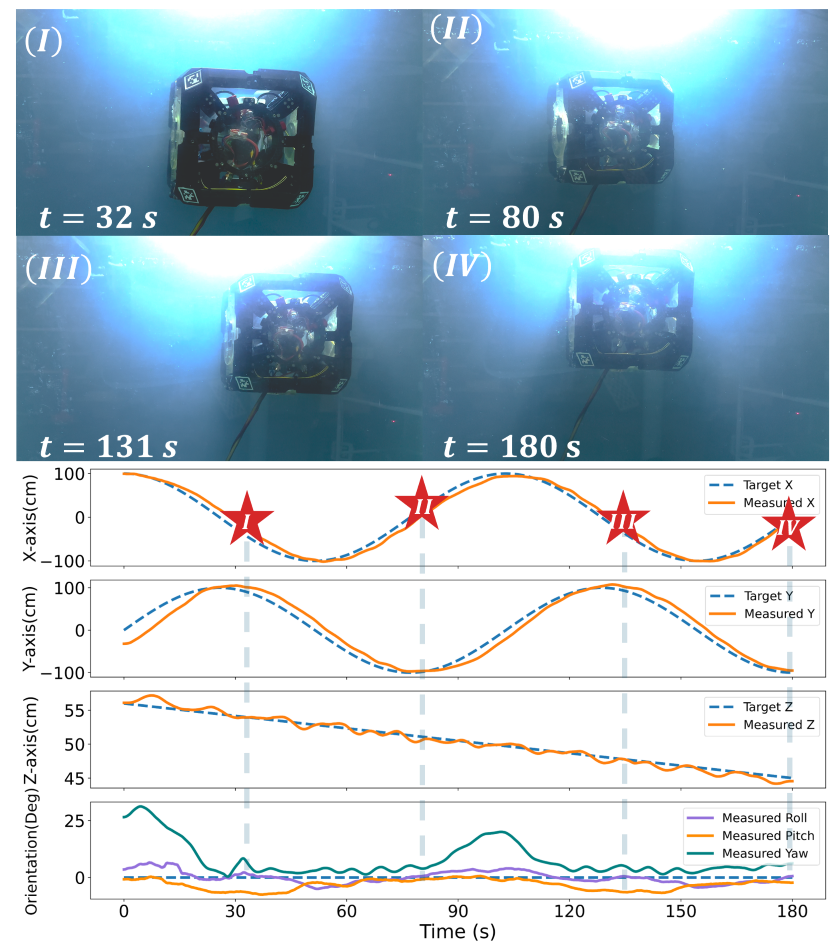}
\vspace{-7pt}
\caption{
Cylindrical spiral trajectory tracking from supplementary video clips S1. \(\textit{I}\)-\(\textit{IV}\) shows snapshots of the tracking process.
Below figure shows the position and orientation of ModCube while tracking the target trajectory. 
Highlighting key events marked by red stars correspond to snapshots.
}
\label{pic:spiral_trajectory_tracking}
\end{figure}

\section{Experiments}
\label{sec:experiments}
\subsection{Experimental Setup}
\subsubsection{Hardware Setup}
We conduct experiments in two water tanks. 
The first is a lab-scale tank with dimensions \(1.3 \times 1.2 \times 0.76 \, \text{m}\), where experiments depicted in Fig.~\ref{pic:spiral_trajectory_tracking} and Fig.~\ref{pic:docking} are performed. 
The second is the CMU Highbay water tank, with dimensions \(\phi 7 \, \text{m} \times 3 \, \text{m}\), used for demo shown in Fig.~\ref{pic:highbay_demo}. 

Both tanks are equipped with an overhead camera system for motion tracking. For all experiments, a \({10}^2\,\text{cm}\) Fiducial Tags is affixed to the top of each ModCube to enable real-time tracking at \(60 \, \text{Hz}\).

Signal transmission for RS-ModCubes is managed via a CANbus converter, serving as the central communication hub. Power is supplied using a \(12 \, \text{V} / 10 \, \text{A}\) DC source. Additionally, a joystick with an emergency stop button facilitated mode switching and manual control.

\subsubsection{Software Setup}
\changed{Our system is developed based on the "Kingfisher" platform~\cite{chemel2020tartan} developed by the TARTANAUV RoboSub Team at Carnegie Mellon University.}

\begin{table}[h!]
\centering
\caption{Self-assembly success rates and failure reasons}
\label{tab:success_rate}
\setlength{\tabcolsep}{4pt} 
\renewcommand{\arraystretch}{1.2} 
\small
\begin{tabular}{lccc}
\toprule
{Category} & {Success} & \multicolumn{2}{c}{{Failure}} \\
\cmidrule{3-4}
& & {Unstable Control} & {Disk Jamming} \\
\midrule
{Sets Number} & 17 & 1 & 2 \\
{Percentage} & 85\% & 5\% & 10\% \\
\bottomrule
\end{tabular}
\end{table}

\subsection{Spiral Trajectory Tracking}
A cylindrical spiral trajectory tracking task is performed with ModCube to validate the accuracy of the proposed dynamics model and control method.
Where ModCube initiates the task at the bottom of the tank and ascends along the predefined spiral path, and holds a hovering orientation, is shown as Fig.~\ref{pic:spiral_trajectory_tracking} ~\(\textit{I}\)–\(\textit{IV}\). 
The tracking results demonstrate the system's capability to follow a target cylindrical spiral trajectory with low position \ac{RMSE} of 20.92 mm.

\begin{figure}[!t]
\centering
\includegraphics[width=0.9\linewidth]{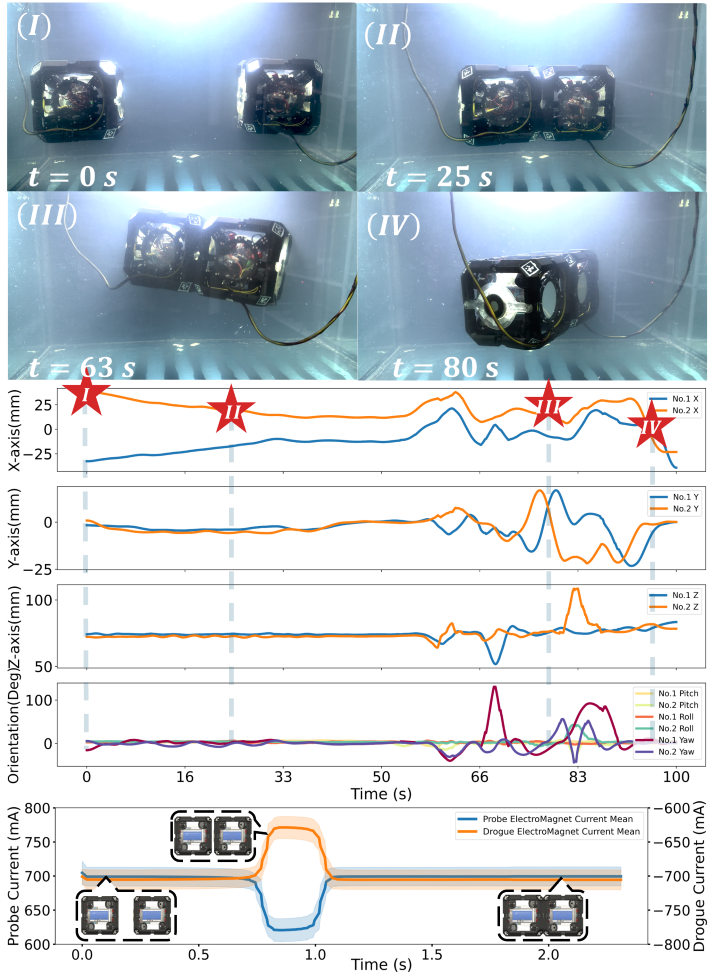}
\vspace{-7pt}
\caption{
Snapshots of the hovering self-assembly of two ModCube modules, and steering experiment of RS-ModCubes from supplementary video clips S2--S3. 
\(\textit{I}\)--\(\textit{IV}\) show the self-assembly process, while \(\textit{III}\)--\(\textit{IV}\) depict RS-ModCubes' teleoperated locomotion using a joystick. 
The mid graph shows the position and orientation throughout the process, with key events marked by red stars corresponding to snapshots. 
The bottom graph presents \ac{MDS} current values during docking, averaged over ten experiments.
}
\label{pic:docking}
\end{figure}

\subsection{Self-assembly}
\label{subsec_assembly}
\changed{Docking commands, issued as trajectory inputs, guide ModCube modules to execute precise docking maneuvers, leveraging their trajectory tracking capabilities.}
Self-assembly capability of RS-ModCubes using two ModCube modules, as shown in Fig.~\ref{pic:docking}. 
In this experiment, two ModCube modules started from opposite sides in lab tank, each following a straight-line trajectory and holding a hovering orientation. 
\changed{Due to experimental constraints, only a single overhead camera was mounted above the water tank, enabling vision-based localization in the vertical direction. This setup limited vertical assembly, as overlapping modules would obscure localization data. As a result, the experiments focused on horizontal self-assembly.}

\changed{A total of 20 experiments, summarized in Table~\ref{tab:success_rate}, were conducted to demonstrate the stability and robustness of proposed model and control methods, achieving a high docking success rate. }

\subsection{Trajectory Planning for RS-ModCubes}
\changed{RS-ModCubes trajectories are planned using the Minimum Snap method~\cite{mellinger2011minimum}, which minimizes the total snap (fourth derivative of position) over a piecewise polynomial path to ensure smooth trajectory generation.
Each segment $k$ of the trajectory is represented as a polynomial:
\(p_k(t) = \sum_{i=0}^n \alpha_{k,i} \, (t - t_{k-1})^i,\)
where $t_{k-1}$ is the start time of segment $k$, $\alpha_{k,i}$ are coefficients, and $n$ is the polynomial order.
The optimization problem defined as:}
\vspace{-5pt}
\changed{\begin{equation}
\begin{aligned}\min  & \quad J = \sum_{k=1}^M \int_{0}^{T_k} \left( \frac{d^4 p_k(t)}{dt^4} \right)^2 dt \\
\text{subject to} & \quad p_k(0) = w_{j}, \quad p_k(T_k) = w_{j+1},  \\
& \frac{d^m p_k(T_k)}{dt^m} = \frac{d^m p_{k+1}(0)}{dt^m}, \quad m \in \{0, 1, 2\}  \\
\end{aligned}
\label{eq:minimum_snap}
\end{equation}}

\begin{figure}[!t]
\centering
\includegraphics[width=0.95\linewidth]{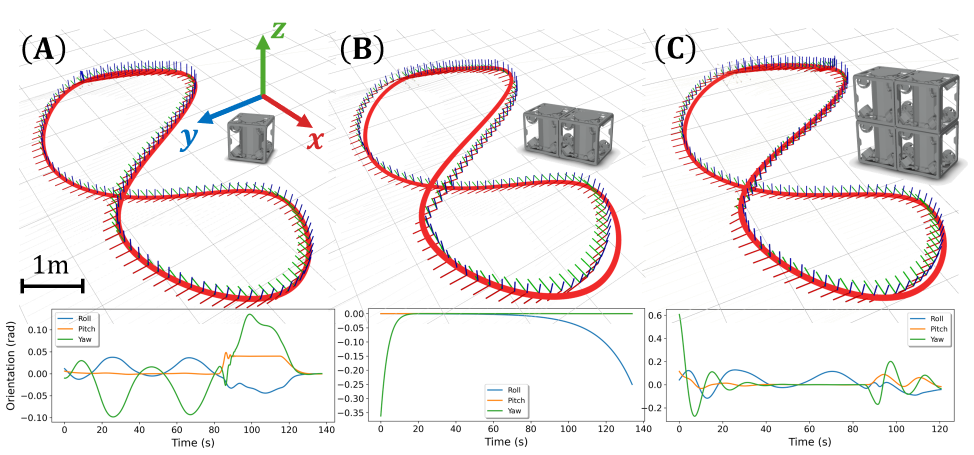}
\vspace{-10pt}
\caption{Results of 3D Möbius strip trajectory tracking experiments with various RS-ModCubes configurations. Trajectories are generated using the minimum snap method~\cite{chemel2020tartan}.  
(A), (B), and (C) illustrate trajectory tracking results for one, two, and four-module configurations of RS-ModCubes, achieving \ac{RMSE} values of 0.4425~cm, 7.6077~cm, and 3.3442~cm, respectively. 
The top row visualizes trajectory tracking in RViz based on Gazebo simulations, while the bottom row presents roll, pitch, and yaw data.}
\label{pic:trajectory_result_rviz}
\end{figure}

\changed{where \( w_j \) and \( w_{j+1} \) are waypoints, and \( T_k \) is the duration of segment \( k \). 
The constraints enforce boundary conditions at waypoints and continuity up to the second derivative (acceleration) between segments.
We utilize the OSQP solver to address the optimization problem. The tracking results are depicted in Fig.~\ref{pic:trajectory_result_rviz}.}

\subsection{Limitations}
\label{Limitations}

\subsubsection{Scalability}
\changed{The system is currently limited to two-module swarm configurations, as increasing the number of modules introduces significant control and coordination challenges.}
\subsubsection{Unstable Assembly and Vertical Assembly Limitations}
\changed{As demonstrated in the supplementary video, instability arises from the low control frequency, constrained by the external motion capture system. Additionally, vertical self-assembly is infeasible due to occlusion issues inherent in vision-based localization, which limits accurate tracking of overlapping modules.}

\section{Conclusion and Future Works}
\changed{This work introduces a reconfigurable and scalable underwater robot system, RS-ModCubes. 
Experimental results demonstrate precise trajectory tracking and stable docking, highlighting robustness in both individual and multi-module configurations, even in constrained environments.
Morphological characterization results, reveal the system's advantages in actuation distribution and energy efficiency compared to four commercial underwater robots, with reconfiguration enabling modulation of dynamic capabilities.}

\changed{Future work will focus on:  
(1) Expanding swarm scalability beyond four modules to enable advanced coordination and complex formations,  
(2) Enhancing control and planning by integrating collision avoidance and more robust vision-based localization for improved precision and adaptability, and  
(3) Developing decentralized, task-specific reconfiguration policies to facilitate large-scale swarm operations and applications.}
\printbibliography
\end{document}